\documentclass[conference]{IEEEtran}
\IEEEoverridecommandlockouts
\usepackage{cite}
\usepackage{amsmath,amssymb,amsfonts}
\usepackage{graphicx}
\usepackage{textcomp}
\usepackage{xcolor}
\usepackage{booktabs}
\def\BibTeX{{\rm B\kern-.05em{\sc i\kern-.025em b}\kern-.08em
T\kern-.1667em\lower.7ex\hbox{E}\kern-.125emX}}

\usepackage{todonotes}
\usepackage{multirow}
\usepackage{caption}
\usepackage{subcaption}
\usepackage{colortbl}
\usepackage{graphicx}
\usepackage{enumitem}

\usepackage{algpseudocode}
\usepackage{xcolor}
\usepackage[linesnumbered,ruled,vlined]{algorithm2e}

\SetCommentSty{mycommfont}

\makeatletter
\newcommand{\linebreakand}{%
  \end{@IEEEauthorhalign}
  \hfill\mbox{}\par
  \mbox{}\hfill\begin{@IEEEauthorhalign}
}
\makeatother

\begin{document}

\title{Amorphous Fortress Online:\\
Collaboratively Designing Open-Ended Multi-Agent AI and Game Environments
\thanks{M Charity was supported by a GAANN Fellowship awarded by the US Department of Education under grant  P200A210096}
}

\author{\IEEEauthorblockN{M Charity}
\IEEEauthorblockA{\textit{Game Innovation Lab} \\
\textit{New York University}\\
Brooklyn, NY \\
mlc761@nyu.edu}

\and
\IEEEauthorblockN{Mayu Wilson}
\IEEEauthorblockA{\textit{Independent Researcher} \\
\textit{Independent}\\
Brooklyn, NY \\
mayuwilson9@gmail.com}
\and
\IEEEauthorblockN{Steven Lee}
\IEEEauthorblockA{\textit{Independent Researcher} \\
\textit{Independent}\\
Brooklyn, NY \\
stevenleesl12345@gmail.com}

\linebreakand
\IEEEauthorblockN{Dipika Rajesh}
\IEEEauthorblockA{\textit{Independent Researcher} \\
\textit{Independent}\\
Chennai, India \\
dipika.rajesh@gmail.com}
\and
\IEEEauthorblockN{Sam Earle}
\IEEEauthorblockA{\textit{Game Innovation Lab} \\
\textit{New York University}\\
Brooklyn, NY \\
smearle@nyu.edu}
\and
\IEEEauthorblockN{Julian Togelius}
\IEEEauthorblockA{\textit{Game Innovation Lab} \\
\textit{New York University}\\
Brooklyn, NY \\
julian@togelius.com}
}

\maketitle

\begin{abstract}
This work introduces Amorphous Fortress Online – a web-based platform where users can design petri-dish-like environments and games consisting of multi-agent AI characters. Users can play, create, and share artificial life (ALIFE) and game environments made up of microscopic but transparent finite-state machine (FSM) agents that interact with each other. The website features multiple interactive editors and accessible settings to view the multi-agent interactions directly from the browser. This system serves to provide a database of thematically diverse AI and game environments that use the emergent behaviors of simple AI agents.

\end{abstract}


\section{Introduction}



Open-endedness in games is difficult to define and harder to design. Open-ended simulation environments and environments that promote artificial life offer a multitude of challenges and emergent scenarios for AI to solve \cite{bedau2000open, stanley2017open}. Such open-ended simulation games include the \textit{Sims} or \textit{Dwarf Fortress} -- where multi-agent interactions provide the core gameloop and create unique game experiences for the player. However, designing AI for these environments is limited to their specific domain and the behavior range of other agents. An AI designed to interact with the world and mechanics of Dwarf Fortress could not interact well with the world and mechanics of the Sims. As such, there exists a limitation to the types of AI that can be designed and trained in these environments and a limitation on the range of their emergent behavior. 

\begin{figure}[!ht]
 \centering
 \begin{subfigure}[h]{0.235\textwidth}
     \includegraphics[width=\textwidth]{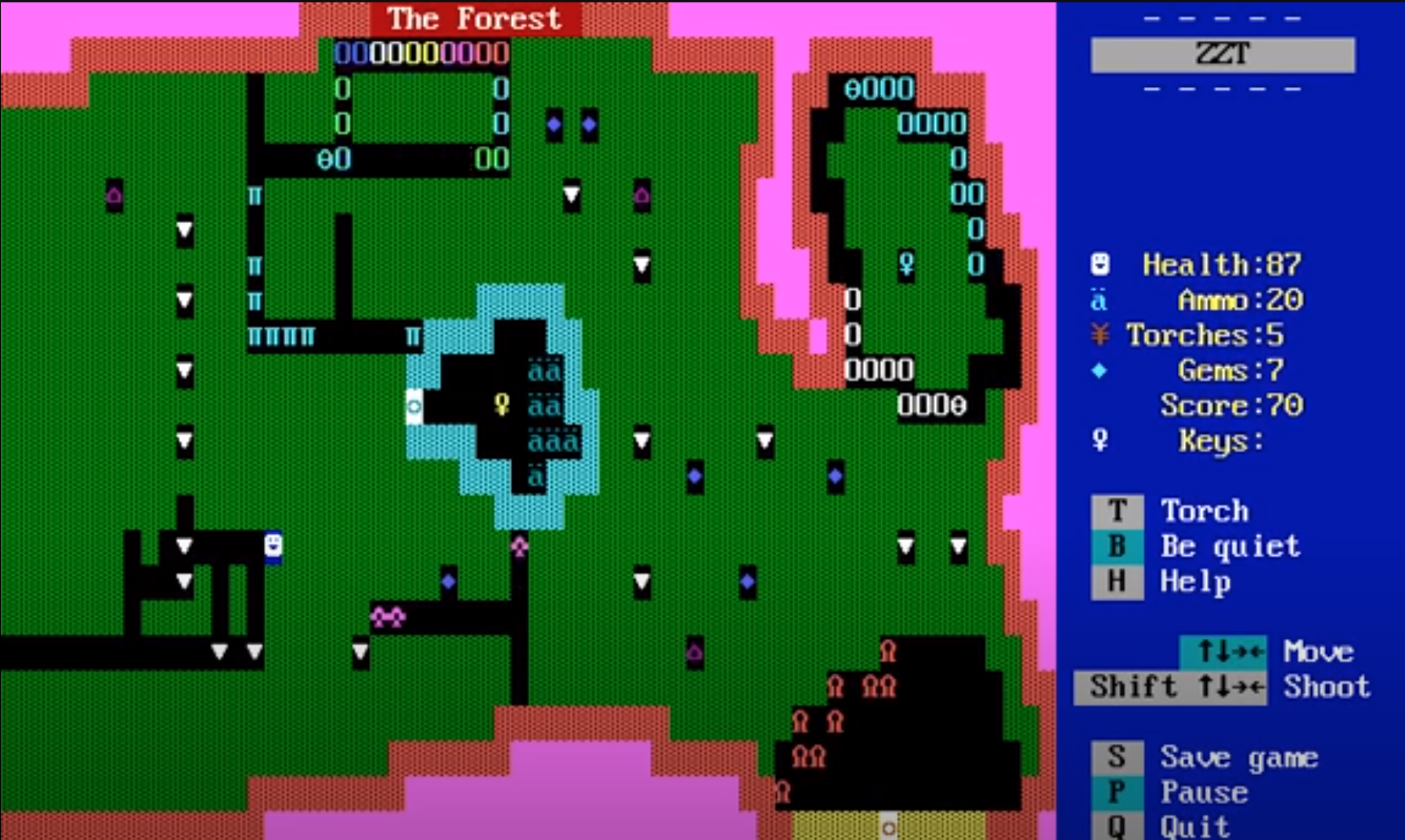}
     \caption{}
 \end{subfigure}
 \begin{subfigure}[h]{0.235\textwidth}
     \includegraphics[width=\textwidth]{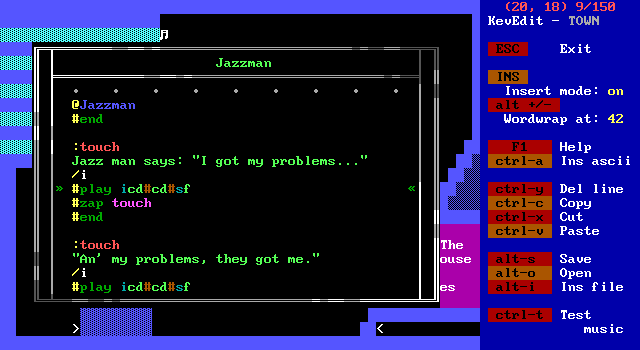}
     \caption{}
 \end{subfigure}
 \begin{subfigure}[h]{0.235\textwidth}
     \centering
     \includegraphics[width=\textwidth]{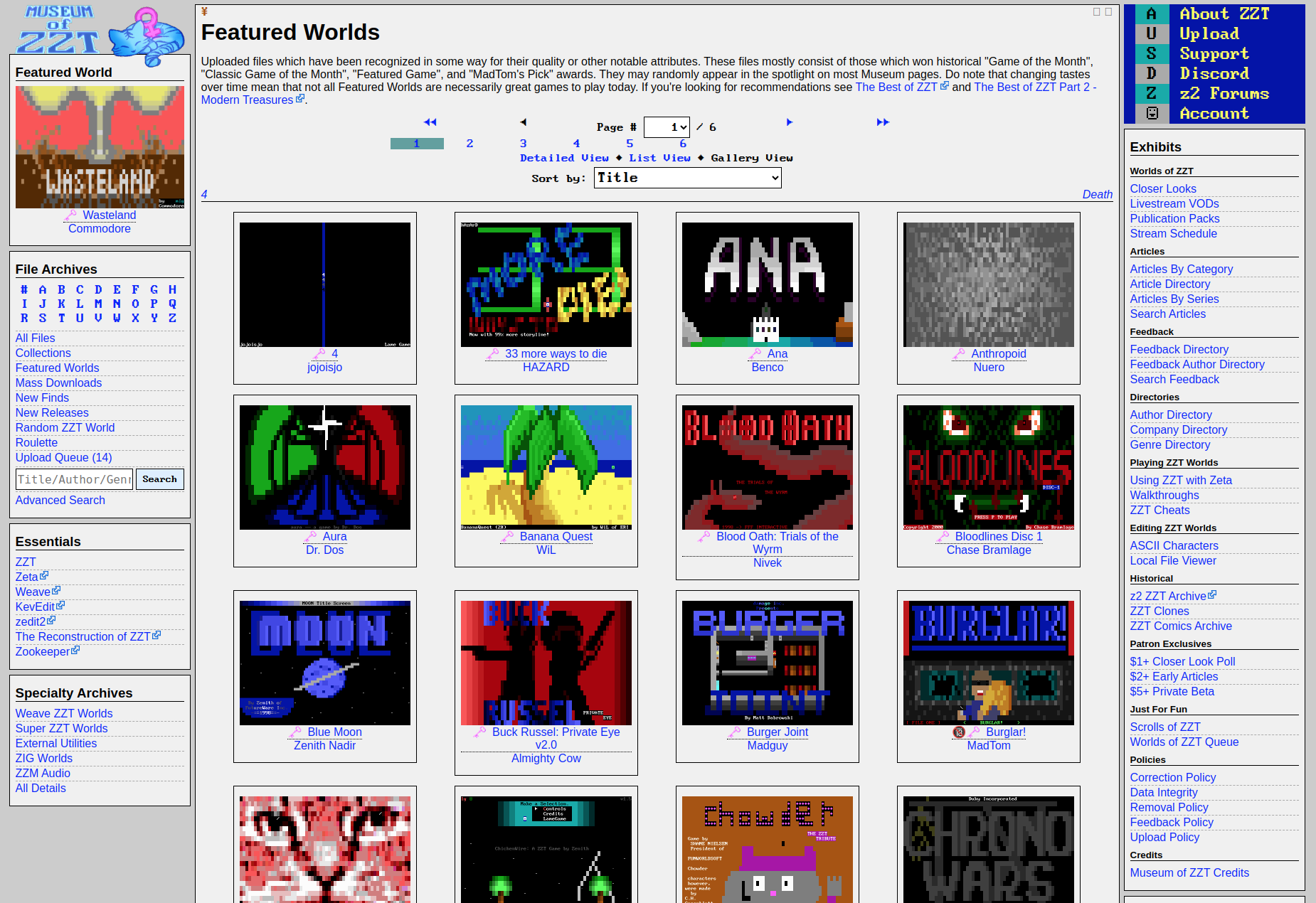}
     \caption{}
 \end{subfigure}
 \begin{subfigure}[h]{0.235\textwidth}
     \centering
     \includegraphics[width=\textwidth]{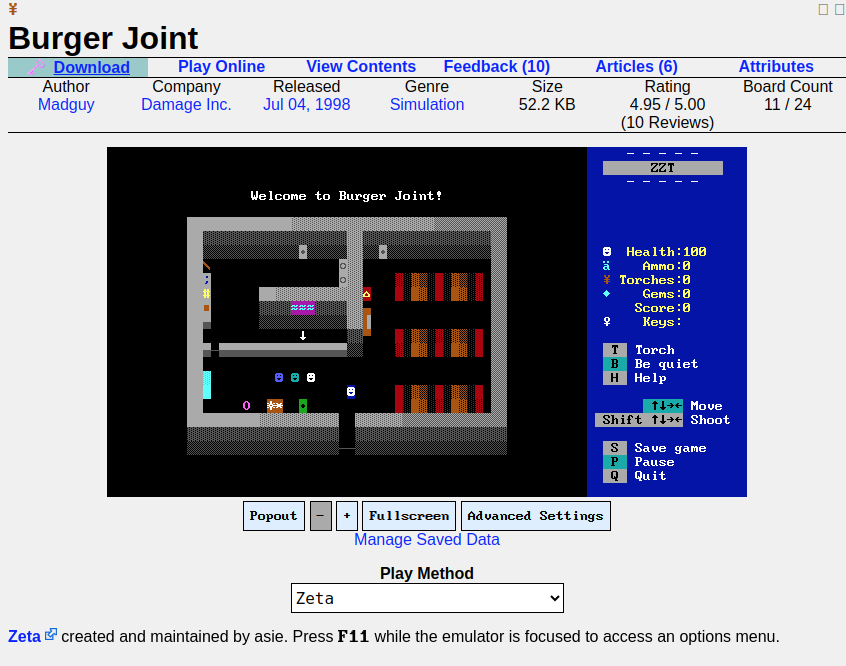}
     \caption{}
 \end{subfigure}
    \caption{The community-based ZZT game design process showing (a) the base action-adventure ZZT game; (b) the ZZT game editor; (c) the Museum of ZZT BBS website page; (d) a user-created simulation role-playing game using the ZZT engine.}
\end{figure}

Community-centric games like \textit{ZZT}\footnote{https://museumofzzt.com/}\footnote{https://museumofzzt.com/file/play/burgerj/}, \textit{Roblox}, and \textit{LittleBigPlanet} rely on the players to design and share the modifiable elements of the game -- such as assets or levels -- to the game's network or database. This creates a constant flow of new and engaging user-generated content for the game. As a result, games with new genres, themes, objectives, and challenges all stem from the same engine and allow players to grow and develop even more content from this shared space. These community games exhibit content with a free-range of creativity to promote open-endedness and undefined boundaries of the environment's genre. By combining the open-endedness of community-centric user-generated content games with the multi-agent aspect of simulation games, we can create a new branch of game AI research that will provide a database for open-ended environment challenges. This database of open-ended AI environments can facilitate emergent behavior and engage players with new AI interactions while providing a much needed diversity of domains and game genres. 

We introduce Amorphous Fortress Online\footnote{http://amorphous-fortress.xyz/} – an online web-based platform where users can design, play, and share small petri-dish-like environments and games consisting of multi-agent AI characters. Users can design artificial life (ALIFE) and game environments made up of microscopic but transparent finite-state machine (FSM) agents. Agents from these environments can be saved and placed in other environments to create new emergent interactions. The website features multiple GUI editors and accessible settings to view the multi-agent interactions directly from the browser. This system serves to provide a database of thematically diverse AI environments and to lay the foundation for future AI work such as training reinforcement learning agents, training recommendation systems, and training generative AI models using the environment and agent definition data submitted by users.


\section{Background}

\subsection{Multi-Agent Simulations}

Simulation games such as the \textit{Sims}, \textit{Dwarf Fortress}, and even \textit{Conway's Game of Life} depend on multiple autonomous agents interacting with each other in order to create unique experiences in the system. Sometimes these interactions lead to emergent behaviors -- behavior unseen or unexpected by the human designers and players for these simulation games. As such, simulation design with multi-agent systems in game AI research can create open-ended challenges through emergent design and unique environments for training AI such as reinforcement learning agents. Earle et. al~\cite{earle2020using} introduce a training environment for reinforcement learning agents in the game \textit{SimCity} and examine population behaviors of cellular automata in \textit{Conway's Game of Life}. Suarez et. al. take the mechanics of a large MMORPG to simulate and discover emergent agent behaviors and interactions in a confined simulation environment \cite{suarez2021neural}. Amorphous Fortress Online uses a multi-agent simulation engine in order to create unique, open-ended experiences and demonstrate emergent behavior from interactions with other autonomous agents and the player, even in such a small confined environment.

\subsection{Community-Made Content Design}

Online game content design and collaborative, community-made content generation exists across multiple genres of games -- typically in the form of levels for games with existing mechanics such as in \textit{Super Mario Maker} or \textit{FreeRider Online} or also in the form of in-game assets such as the texture pattern sharing feature of the Animal Crossing series. With an online sharing system for a game, a community of players can form around designing unique content for a game that others can use. In AI research, user created content -- generated with or without AI assistance -- can help to train and design AI systems capable of generating unique content. This design feedback loop between AI model and user can encourage more user engagement with the system and guide the generative design process. Examples of community-made content being used with AI systems include Picbreeder, an online system where users can guide the generation process of AI made images through interactive evolution and submit them for others to evolve on the website \cite{secretan2008picbreeder}. Similarly in game design, Baba is Y'all's recommendation system learns from user submitted puzzle levels for the Baba is You game to make levels with new mechanic combinations not previously found in the database \cite{charity2020baba}. The Aesthetic Bot system utilizes the crowd-sourced voting system of the Twitter API to compare the aesthetic quality of game maps against user submitted maps in order to improve its subjective map generation quality \cite{charity2022aesthetic}. These systems demonstrate how creating a community that facilitates a constant flow of content generation and feedback reporting can be effective methods of guiding the generative process of AI systems.

\subsection{Open-Ended Game Design}

Open-endedness and creativity in game design also makes use of community-made content generation to engage players with  the system. However, with open-ended games, there is no limit to the domain or genre. The custom object-oriented programming language and bulletin board system (BBS) of the shareware game \textit{ZZT} is an example of how players can remix and design the grid-world and ASCII characters to make new games including puzzle games, RPGs, and visual novels \cite{anthropy2014zzt}. \textit{LittleBigPlanet} is a platformer game designed around crowd-sourced content generation where users can share asset creations and upload their own levels using the game's base mechanics\cite{lbp1}. The sequel game, \textit{LittleBigPlanet2}, expanded the genre domains outside of platformers including racing games, arcade multiplayer games, and even movies\cite{lbp2}. This later iteration of the series added more complex mechanics such as simple AI behavior programming through the Sockbots and logic gate programming for switches and other mechanisms.

In the areas of game AI research and the independent game development, much work has been done to develop game engines that facilitate online communities. User-bases for micro-game engines and fantasy console engines (i.e. Bitsy, PICO-8) allow users to share games made with these specific engines on platforms such as Itch.io or the website's BBS. The accessibility and simplicity of these game engines are what make them so popular and distinguishable in their design \cite{warren2019tiny}. The ANGELINA system by Cook et. al. uses automated game design principles to design new games across multiple genres\cite{cook2016angelina1,cook2016angelina2}. An Itch page\footnote{https://gamesbyangelina.itch.io/} around these generated games has been made to document the works output by the system. With Amorphous Fortress Online, we would like the system to act as an open-ended game engine and interface to allow users to create a community of new AI environments without limits on the domain or genre.


\section{Amorphous Fortress Online}

Amorphous Fortress Online is an open-ended simulation and game design system that allows users to create small multi-agent environments. These environments, or \textit{fortresses} are made up of AI agents called \textit{entities} that interact with each other in a simulation. Users can also make playable character entities that also interact with the automated entities in the fortress -- turning the zero-player simulation into a user controlled game environment. The following sections detail the components of the online system and the design pipeline users undergo to create, remix, and play the fortresses submitted to the website.

\subsection{Amorphous Fortress Engine}

The underlying engine of the Amorphous Fortress Online system is a JavaScript-ported version of the Amorphous Fortress Python engine \cite{charity2023amorphous,earle2023quality}. Many elements from the Python engine are retained in the JavaScript port, however Amorphous Fortress Online has further additions and improvements for deeper agent interaction and gameplay in the fortresses.

\begin{figure}[!h]
    \centering
    \includegraphics[width=\linewidth]{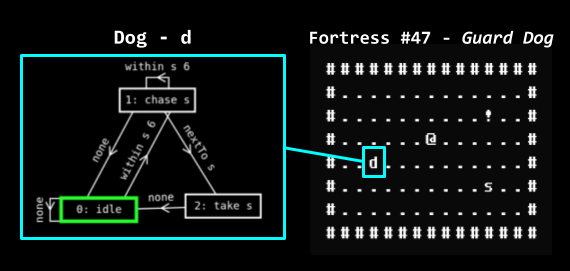}
    \caption{A visualization of an entity's behavior FSM structure and an instance of its class in the fortress environment}
    \label{fig:fsm_fort}
\end{figure}

Each entity class of the Amorphous Fortress is defined by a singular ASCII character, a list of nodes, and a set of edges. Each instance of an entity has a unique identifier value. The ASCII character-rendering was directly inspired by classic terminal-based rogue-likes such as Rogue and Dwarf Fortress. These characters can be any symbol that can be represented on a terminal but lack the color variance found in some text-based roguelike games. Future iterations of Amorphous Fortress Online will introduce colors and sprite designs for better visual engagement.

The behavior for the autonomous entities of the system are defined using Finite State Machines (FSMs) -- abstract representation of an interconnected set of actions, states, and transitions represented by a graph that have been used to control a multitude of simple game AI behaviors \cite{georgios2018artificial}. The nodes of the FSM graph define the action of the agent while the edges determine conditions for transitioning between action states. The simplicity, explainability, and scalability of the FSM design creates an accessible interface for users to design new interactive AI behaviors in the Amorphous Fortress engine.

There are a total of 13 action behavior nodes defined for the entities' FSMs in the Amorphous Fortress. 10 of these nodes define automated behavior for the entity; actions that will occur without user interaction in the simulation. These nodes are defined in Table \ref{tab:auto_nodes} and based on common AI game mechanics such as \textit{chase}, or \textit{add}. The remaining three action behavior nodes are player nodes -- nodes that depend on user input to enact the action. These three nodes $player\_move$, $player\_push$, and $player\_move\_wall$ are based on their autonomous action nodes $move$, $push$, $move\_wall$ respectively. However, instead of using random movements, they use 5 directional user inputs: up, down, left, right, and a skip turn action which keeps the entity idle. Like the original Amorphous Fortress engine, each node can only be used once (excluding character permutations) to avoid redundancy.

\begin{table}[!th]
    \begin{tabular}{p{0.19\linewidth}  p{0.73\linewidth}}
    \toprule
        Action Node   & Definition                  \\
        \hline
        \midrule
        idle          & \textit{the entity remains stationary}  \\
        move          & \textit{the item moves in a random direction (north, south, east, or west)}              \\
        die           & \textit{the entity is deleted from the fortress}      \\
        clone         & \textit{the entity creates another instance of its own class}    \\
        push (c)      & \textit{the entity will attempt to move in a random direction and will push an entity of the specified target character into the next space over (if possible)}  \\
        take (c)      & \textit{the entity removes the nearest entity of the specified target character}     \\
        chase (c)     & \textit{the entity will move towards the position of the nearest entity of the specified target character}     \\
        add (c)       & \textit{the entity creates another instance from the class of the specified target character}      \\
        transform (c) & \textit{the entity will change to a different entity class - thus altering its FSM definition entirely}        \\
        move\_wall (c)      & \textit{the entity will attempt to move in a random direction unless there is an entity of the specified class at that position - otherwise it will remain idle}
    \end{tabular}
    \caption{Entity FSM action autonomous node definitions}
    \label{tab:auto_nodes}
\end{table}

The conditional edges define the transitions of the entity from one action state to another.  A node for an entity can have multiple edges but only two directional edges can exist between two nodes (i.e. $A \rightarrow B$ and $B \rightarrow A$). The edges of a currently active node will transition to their corresponding node based on edge value priority. These conditions are defined in Table \ref{tab:condition_edge} in order from lowest to highest priority.

\begin{table}[!th]
    \begin{tabular}{p{0.18\linewidth}  p{0.75\linewidth}}
    \toprule
        Condition   & Definition                  \\
        \hline
        \midrule
        none          & \textit{no condition is required to transition states}  \\
        step (\#)    & \textit{every \# number of simulation ticks the edge is activated and the node transitions}              \\
        within (c) (\#)        & \textit{checks whether the entity is within \# spaces from an instance of another entity with the target character $c$}       \\
        nextTo (c)        & \textit{checks whether the entity is within one space (north, south, east, or west) of another entity of the target character $c$}    \\
        touch (c)      & \textit{checks whether the entity is in the same space as another entity of the target character $c$}  \\
    \end{tabular}
    \caption{Entity FSM conditional edge definitions (ordered by least to greatest priority)}
    \label{tab:condition_edge}
\end{table}

The \textit{fortress} contains the environment where the simulation takes place and stores general information accessible to all of the entities defined. Each fortress in Amorphous Fortress Online consists of a 16 x 8 grid space containing all of the active entity instances in the simulation, with walls \# as a border and floors $.$ inside. The fortress has 3 termination conditions that can occur that are checked by the engine and are defined as follows:
\begin{itemize}
    \item \textit{extinction}: checks whether there are no entities at all left in the simulation
    \item \textit{overpopulation}: checks whether there exist more than 168 ($2(wh)$) entities in the simulation
    \item \textit{inactivity}: checks whether an action has been taken by an entity in the last 100 timesteps
\end{itemize}

Figure \ref{fig:fsm_fort} shows an example of the FSM behavior definition for the entity class ``Dog'' represented by the character \textit{d} in the submitted fortress called ``Guard Dog'' with a database ID of 47.

The engine for the fortress maintains the simulation loop and states of each entity. Each entity is evaluated at its current node and conditions are checked from the edges to move from that node onto a connecting node if possible. The engine continues updating the simulation until any of the termination conditions have been reached.

\subsection{Example Scenario}

\begin{figure}[!h]
    \centering
    \includegraphics[width=\linewidth]{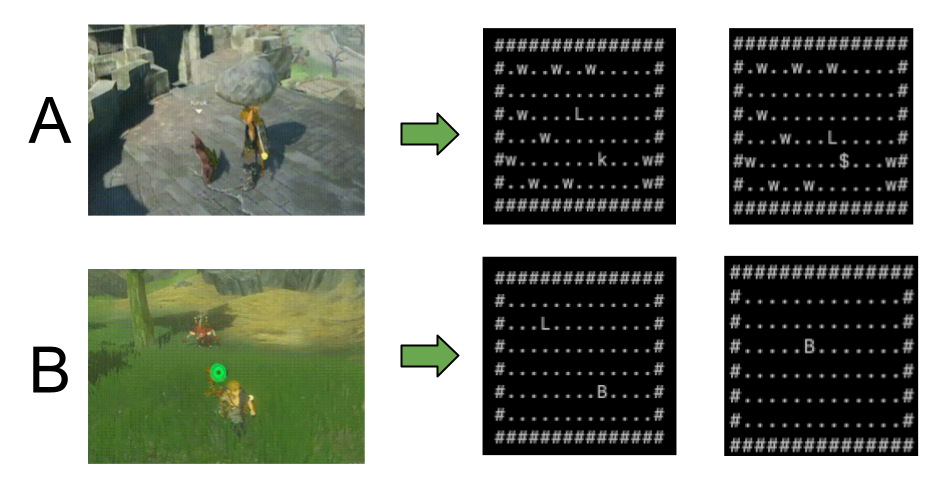}
    \caption{Two example agent interactions from Zelda: Tears of the Kingdom remade in the Amorphous Fortress framework}
    \label{fig:zelda-ex}
\end{figure}

To illustrate the potential of our system to emulate abstracted game and simulation dynamics, we use the FSM entity class definitions to design two simple example environments. These environments, inspired by \textit{The Legend of Zelda: Tears of the Kingdom}~\cite{zelda_tok}, contains three entity class definitions: `Link', `Bokoblin' and `Korok'. In the Zelda franchise, Koroks are characters normally revealed after the player, controlling the character Link, completes a puzzle or interacts with a unique hidden object, at which point the Korok gifts the player a seed that can be used as in-game currency. Bokoblin characters in the game also interact with the character Link by chasing him when he comes into range and trying to attack him. Figure \ref{fig:zelda-ex} shows a screenshot of both scenarios in game and recreated in the framework.

For scenario A, a single Link entity $L$ and a single Korok entity $k$ are initialized in the fortress with grass entities $w$ placed as decoration. Link's FSM is defined such that the character will \textit{move} randomly on the map every \textit{2 steps}. When the Korok (`k') is \textit{next to} Link (`L'), it \textit{transforms} into a seed, `\$'. When Link \textit{touches} the seed, he \textit{takes} it, removing the `\$' character from the map. For scenario B, a Bokoblin entity $B$ will \textit{move} randomly in the fortress. If Link is \textit{within $5$} spaces of the Bokoblin, the Bokoblin will \textit{chase} him. If the Bokoblin is \textit{touching} Link, Link will \textit{die}.

\subsection{Entity Editor}

\begin{figure}[!h]
    \centering
    \includegraphics[width=\linewidth]{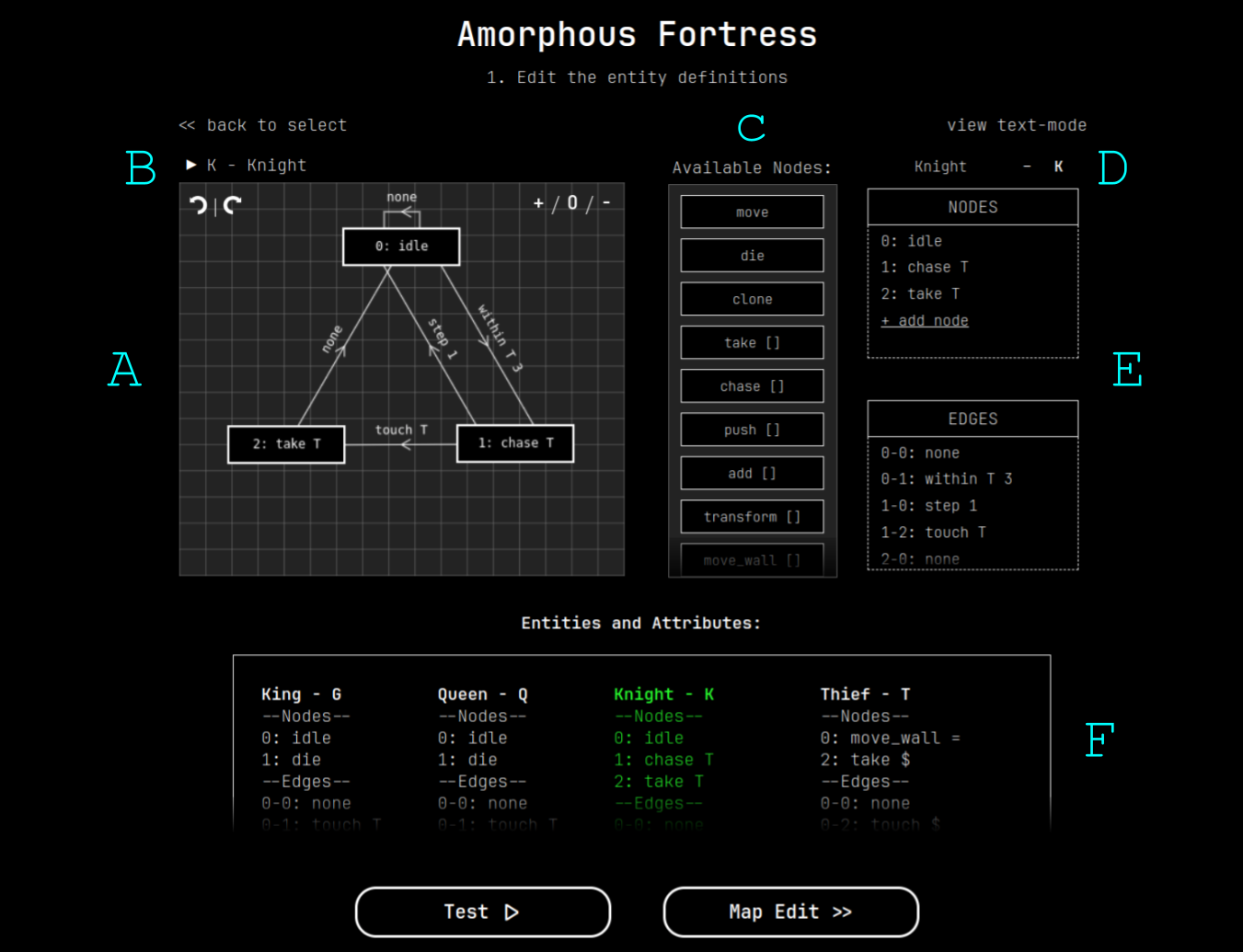}
    \caption{The entity editor screen for Amorphous Fortress Online labeled as follows: A) FSM canvas visual editor; B) Entity selection dropdown; C) Canvas node list; D) Name and character input area; E) Entity node and edge list edit area; F) Fortress entity list}
    \label{fig:ent_editor}
\end{figure}

When designing a new fortress or editing a previously submitted fortress in Amorphous Fortress Online, users are initially taken to the entity editor screen -- referenced in Figure \ref{fig:ent_editor}. Entities are designed and edited with an interactive drag-and-drop and dropdown menu system. The nodes and edges of an entity's FSM are visualized and organized in a canvas editor (A). Users can add new nodes to the canvas by dragging nodes from the available node list section (C.) In the canvas, users can edit a node's value or delete it by right clicking on its FSM box and selecting through its dropdown menu screen. Edges can be added by selecting the starting node box and then clicking on its connecting end node. The edge condition can also be changed in a similar fashion as the nodes by right clicking on it. Alternatively, users can use list editor menu (E) to make changes directly based on value and add and delete entity FSM elements. 

The user can quickly switch between the editing the current entity definitions through a dropdown menu above the visual FSM editor or by clicking on the entity's full definition box from the entity list area (F.). Users can input whatever name or character representation they would like their entity to have in the fortress (D.) Entities in the fortress are added or deleted with a dropdown menu (B) above the canvas editor. 

Each drop down menu has a limited selection of options for the user to choose from when designing their entity. The options for the node and edge values are limited to only the characters that have already been defined in fortress. For example, in a fortress with the entity character representations \textit{M, \&, and +} for the action node \textit{push} only the options \textit{push M}, \textit{\&} and \textit{push +} would be shown.

\subsection{Fortress Editor}

\begin{figure}[!h]
    \centering
    \includegraphics[width=\linewidth]{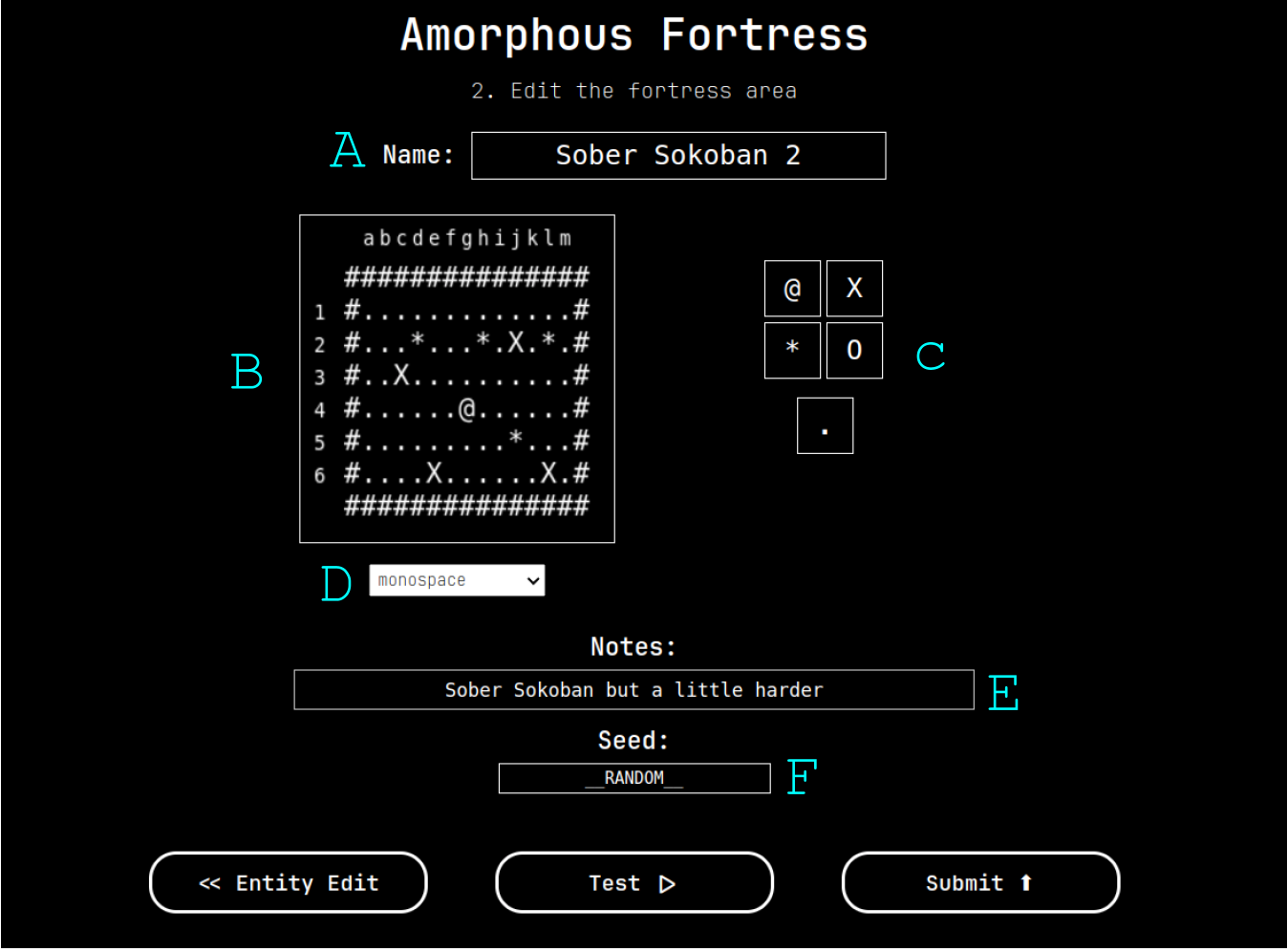}
    \caption{The fortress editor screen for Amorphous Fortress Online labeled as follows: A) Fortress name input area; B) Fortress placement canvas C) Entity palette; D) Font selection; E) Notes input area; F) Seed input area}
    \label{fig:fort_editor}
\end{figure}

After editing their entities, users can use the fortress editor to then place them in the fortress environment -- referenced in Figure \ref{fig:fort_editor}. The fortress editor features a canvas window (B) with the fortress layout in a 14 by 6 grid space area. Users select the entity based on their character representation from a palette (C) on the right and place them onto the fortress map on the left of the screen. The font of the characters and fortress ASCII representation can be changed to 6 different options through a dropdown selection (D.) The fortress editor also includes input areas for users to name their fortress (A), provide notes or a description about the fortress (E), and specify a seed (F) to initialize the fortress with when a user plays it. If the seed is specified with the value ``\textit{\_\_RANDOM\_\_}'' then the fortress will initialize with a random seed every time the simulation is reset.

\subsection{Text Editor}

\begin{figure}[!h]
    \centering
    \includegraphics[width=\linewidth]{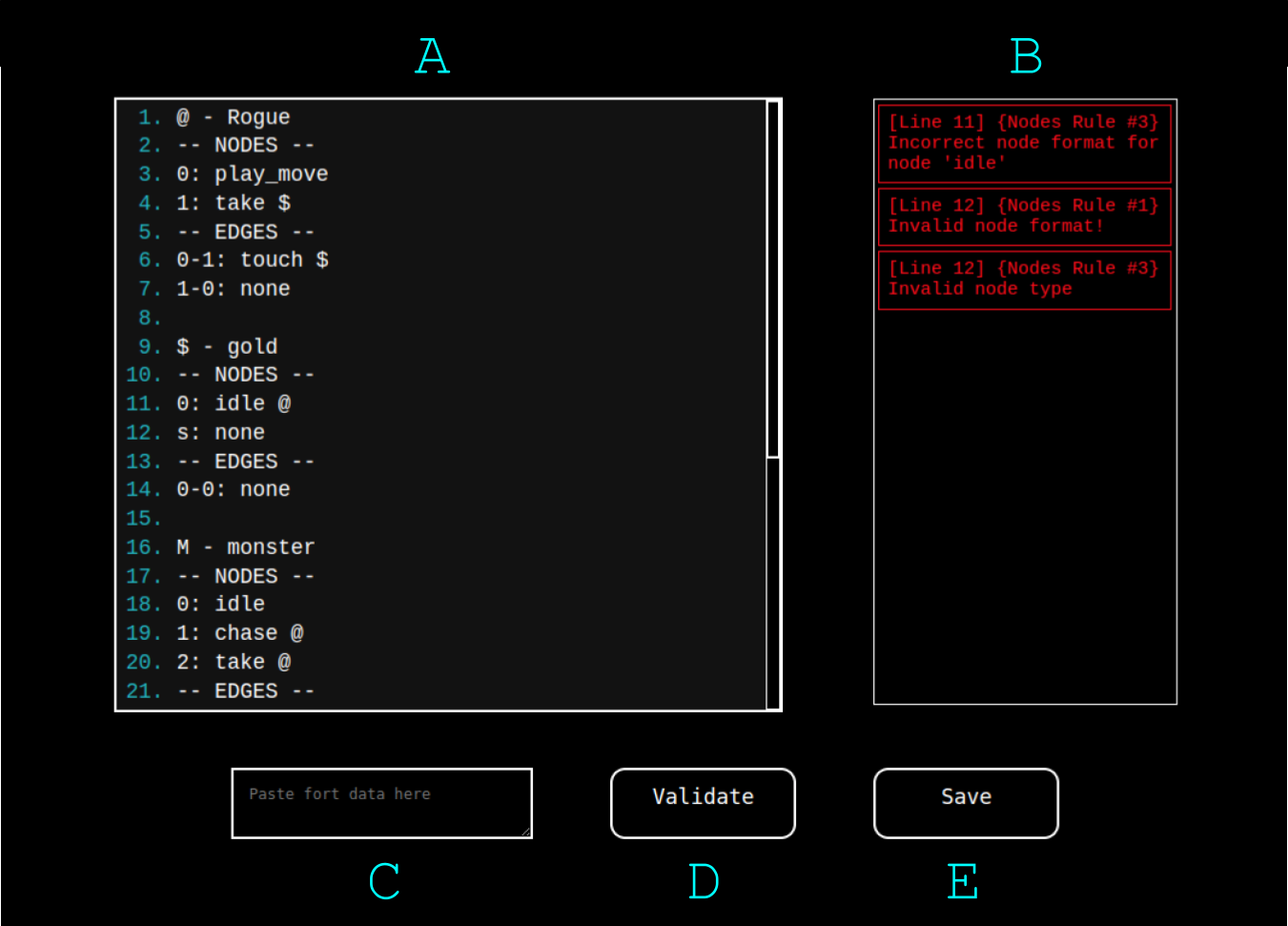}
    \caption{The text editor screen for Amorphous Fortress Online labeled as follows: A) Fortress text editor area; B) Compiler console; C) Raw text paste area; D) Validation check button; E) Save fortress button}
    \label{fig:text_editor}
\end{figure}

For advanced users, Amorphous Fortress Online also includes a text editor -- see Figure \ref{fig:text_editor} -- accessible from the entity editor screen to allow users to directly edit the fortress definition. The text editor acts as an alternative to both the interactive entity and fortress editors. This particular editor is based off of the original Amorphous Fortress fortress definition files. Users can directly specify the entity FSM definitions and initial fortress layout with the text area interface (A.) The text editor features a custom made compiler for the Amorphous Fortress engine that checks for invalid nodes or edges, such as unknown node definitions, invalid FSM parameters, invalid characters, or invalid fortress layouts such as unknown characters or incorrect fortress dimensions. To save the fortress (E) to the entity and fortress GUI editors, users must first validate (D) their input text through the compiler. This is to ensure that the text version of the fortress definition can be fully interpreted by the engine without issue. The compiler returns the type of error found from the text editor as well as the line number in a console (B) to the right of the editor for easier debugging for the user. Users can also paste entire fortress definitions into a smaller text area (C) and edit it line by line. 

\subsection{Play Screen}

\begin{figure}[!h]
    \centering
    \includegraphics[width=\linewidth]{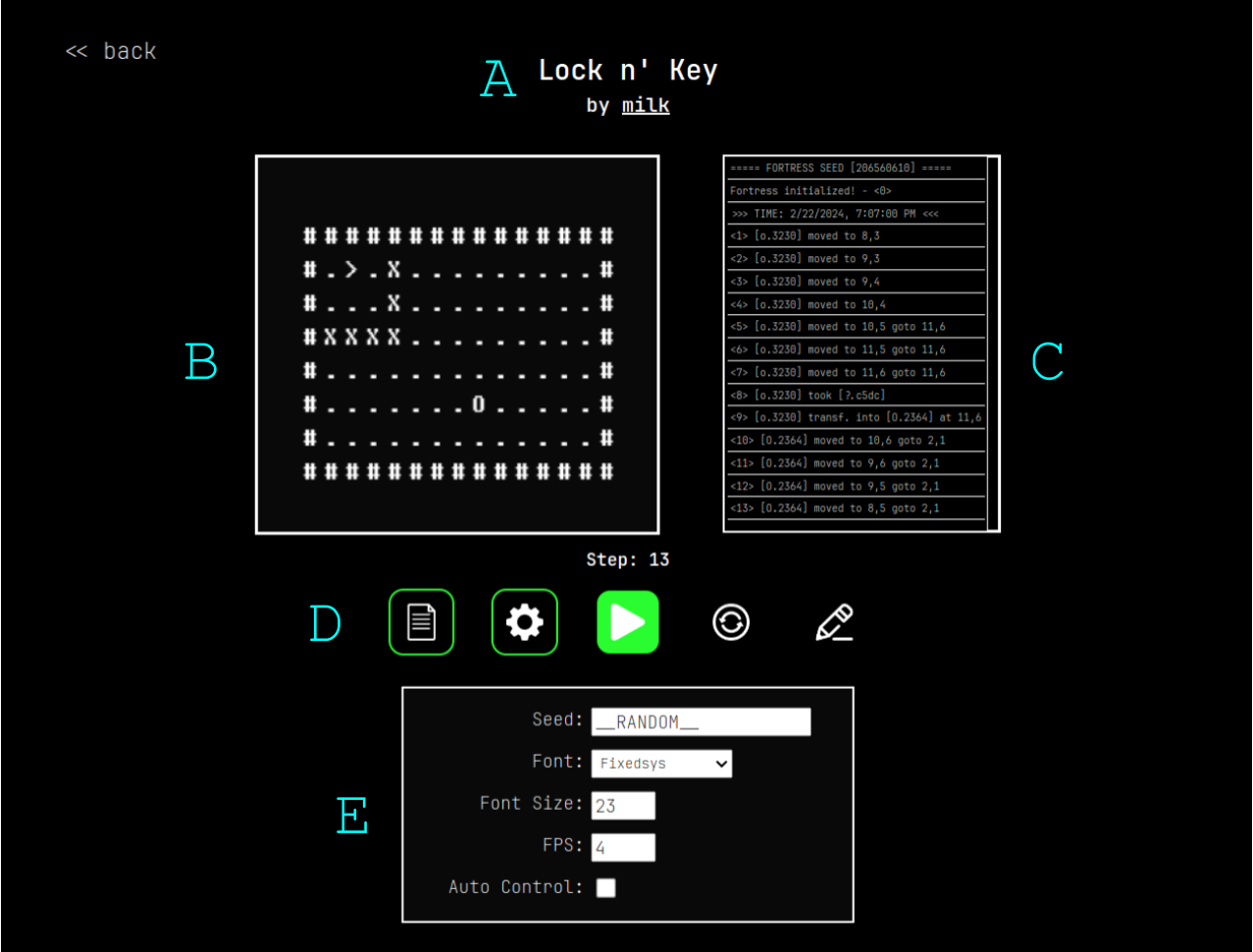}
    \caption{The fortress play screen for Amorphous Fortress Online labeled as follows: A) Name and author of the fortress; B) Main fortress view area; C) Fortress simulation log; D) Simulation menu bar; E) Play screen settings area}
    \label{fig:play_screen}
\end{figure}

From both the entity editor and fortress editor screen, users can test and simulate their fortresses before submitting to the Amorphous Fortress Online database in the play screen (shown in Figure \ref{fig:play_screen}). The play screen details the name and author of the current fortress (A) and shows the fortress state during simulation in the main fortress view window (B) starting with its initial entity placement state. In the menu bar (D) under the main fortress view window, users can start, pause, or reset the simulation by pressing the play button or reset buttons. The play screen also includes a settings area (E) with many user adjustable options for altering the display or the fortress's multi-agent simulation settings. Here, the seed for random agent behavior can be modified to recreate specific agent interactions that occur in the fortress. The font type -- with 6 different font options -- and size of the font displaying the characters in the fortress can be changed by the user. The speed of the simulation can also be adjusted in this settings area. To the right of the fortress view window, the user can display the log of the simulation (C) maintained by the fortress engine. Here every interaction that occurs with the entities is recorded at each timestep. 

For fortresses that include player action nodes, the user can control entities highlighted with a cyan background and specify their input directions with 3 different keyboard control schemes (see Table \ref{tab:control_scheme}). Players can also start and stop the simulation with the P key and reset the simulation with the R key. The settings area features a toggle option to turn off player control and will instead default the movement of player controlled entities in the simulation to automated random directional key presses.

\begin{table}[!th]
    \centering
    \begin{tabular}{llllll}
    \rowcolor[HTML]{C0C0C0} 
    Control Scheme & Up & Down & Left & Right & Skip \\
    Arrow Keys & $\uparrow$ & $\downarrow$ & $\leftarrow$ & $\rightarrow$ & Space \\
    WASD           & W  & S    & A    & D     & F    \\
    Vi Keys        & K  & J    & H    & L     & ;   
    \end{tabular}
    \caption{Control scheme options for player controlled entities}
    \label{tab:control_scheme}
\end{table}

\subsubsection{X-Ray Feature}

\begin{figure}[!h]
    \centering
    \includegraphics[width=\linewidth]{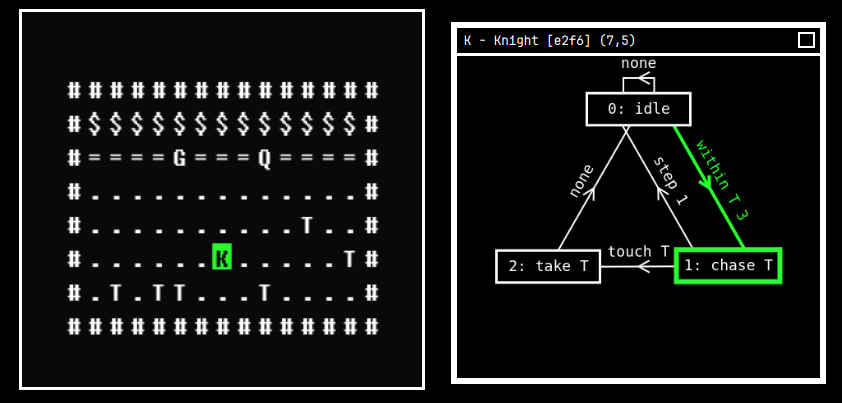}
    \caption{The X-Ray feature (right) highlighting the Knight entity \textit{K} (left) in the Thief Castle fortress (\#36.) The green outlined FSM components indicate active or recently activated behavior logic. Currently its \textit{1: chase T} node is active and because of the edge transtition from the \textit{0-1: within 3 T} edge.}
    \label{fig:xray_window}
\end{figure}

The play screen features a pop-up and movable X-ray window (see Figure \ref{fig:xray_window}) to view an entity's current FSM state during runtime. This window shows the entity's FSM definition as organized by the user in the entity editor screen which is saved with the fortress data upon submission. Active nodes and most recently transtitioned edges are highlighted in green in the X-Ray window and it is constantly updated with each step in the simulation. Entities currently being viewed in the X-ray appear highlighted in green in the main view window. This feature allows for more explainability with the agent behaviors and helps designers debug their agents during fortress creation. 

\subsection{Main Page}

\begin{figure*}[!ht]
     \centering
     \begin{subfigure}[b]{0.5\linewidth}
         \centering
         \includegraphics[width=\linewidth]{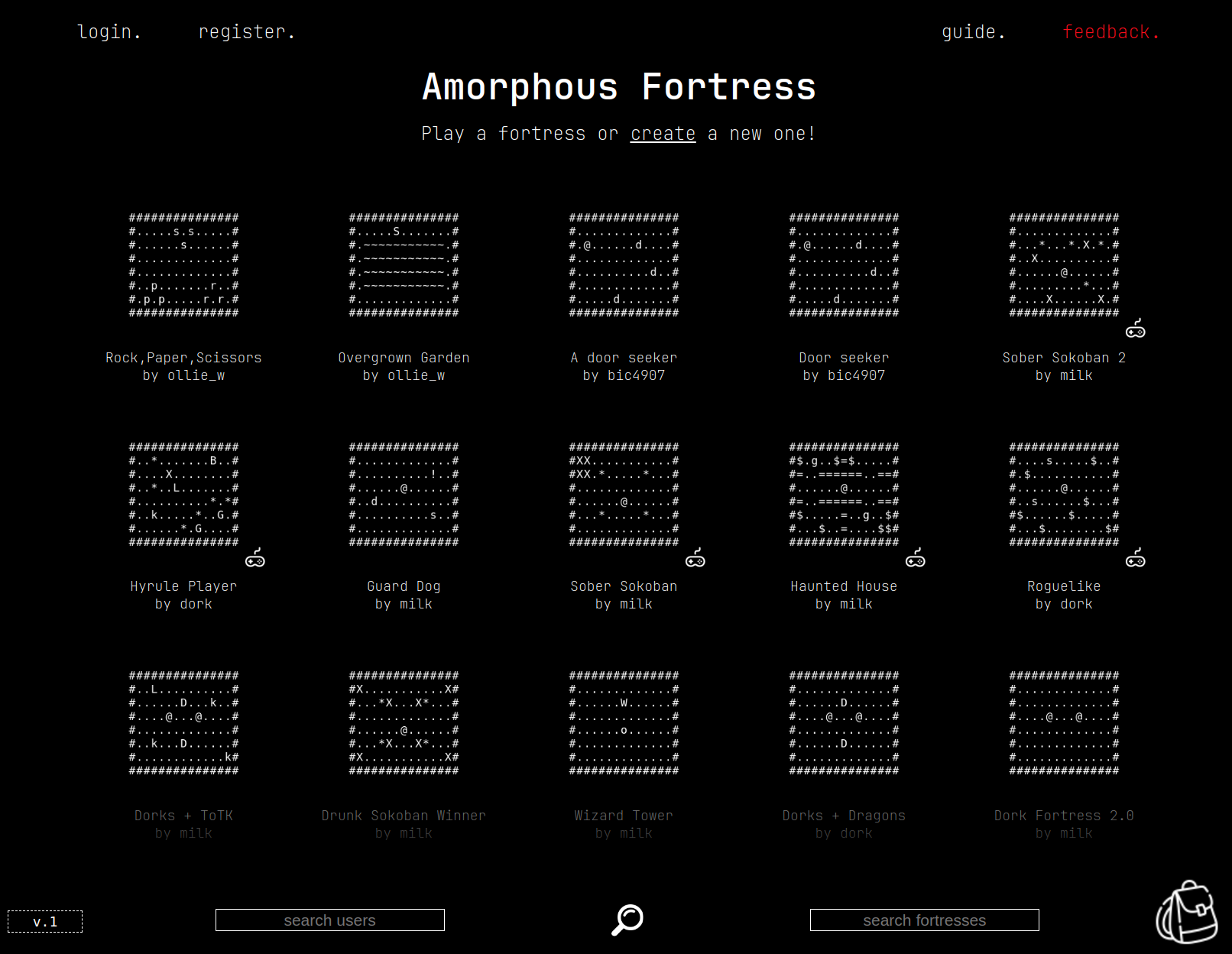}
         \caption{Main screen fortress selection}
         \label{fig:main_matrix}
     \end{subfigure}
     \hfill
     \begin{subfigure}[b]{0.47\linewidth}
         \centering
         \includegraphics[width=\linewidth]{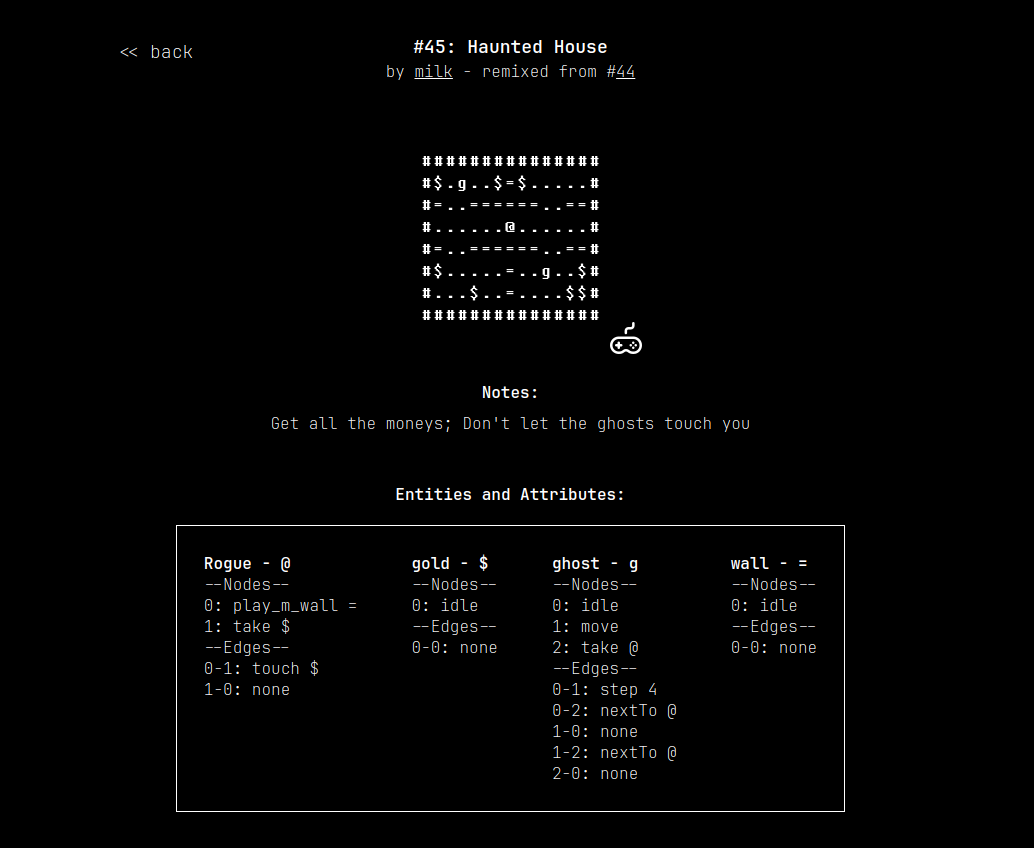}
         \caption{Fortress info viewer}
         \label{fig:main_view}
     \end{subfigure}
        \caption{Main screen of Amorphous Fortress Online with fortress selector and viewer}
        \label{fig:tmain_screen}
\end{figure*}

Once a user finishes editing their fortress, it is uploaded from the fortress editor to the Amorphous Fortress Online database and shown on the main landing page of the website (Figure \ref{fig:main_matrix}). The main page shows the most 120 recent fortresses submitted to the database with their fortress name, author, and initial fortress layout. Hovering over a fortress in the matrix list allows users to see the defined entity names in the fortress. Fortresses that include player action nodes include an outlined game controller next to their fortress rendering to indicate to users that they can be user controlled. 

When a user clicks on the fortress, they can view the full information, including the entity definitions, notes description about the fortress, and a link to the parent fortress ID if the fortress was remixed and edited by the author from a previously submitted fortress. An example fortress view is shown in Figure \ref{fig:main_view}. From this information view, users can either select to play the fortress themselves in the play screen window or edit and remix the fortress to make a new submission.

The main screen features two search bars at the bottom of the page to allow users to find specific fortresses submitted to the database. Users can search for fortresses by username, fortress name, or both. The most recently submitted fortresses matching the search criteria are returned. Users can register a username in order to claim authorship to their fortresses with a login authentication system. For password recovery, there is also an optional input to provide an email address with the username registration. Fortresses submitted anonymously without a user login will be authored by the default "dork" user account.

\subsubsection{Backpack Feature}

\begin{figure}[!h]
    \centering
    \includegraphics[width=\linewidth]{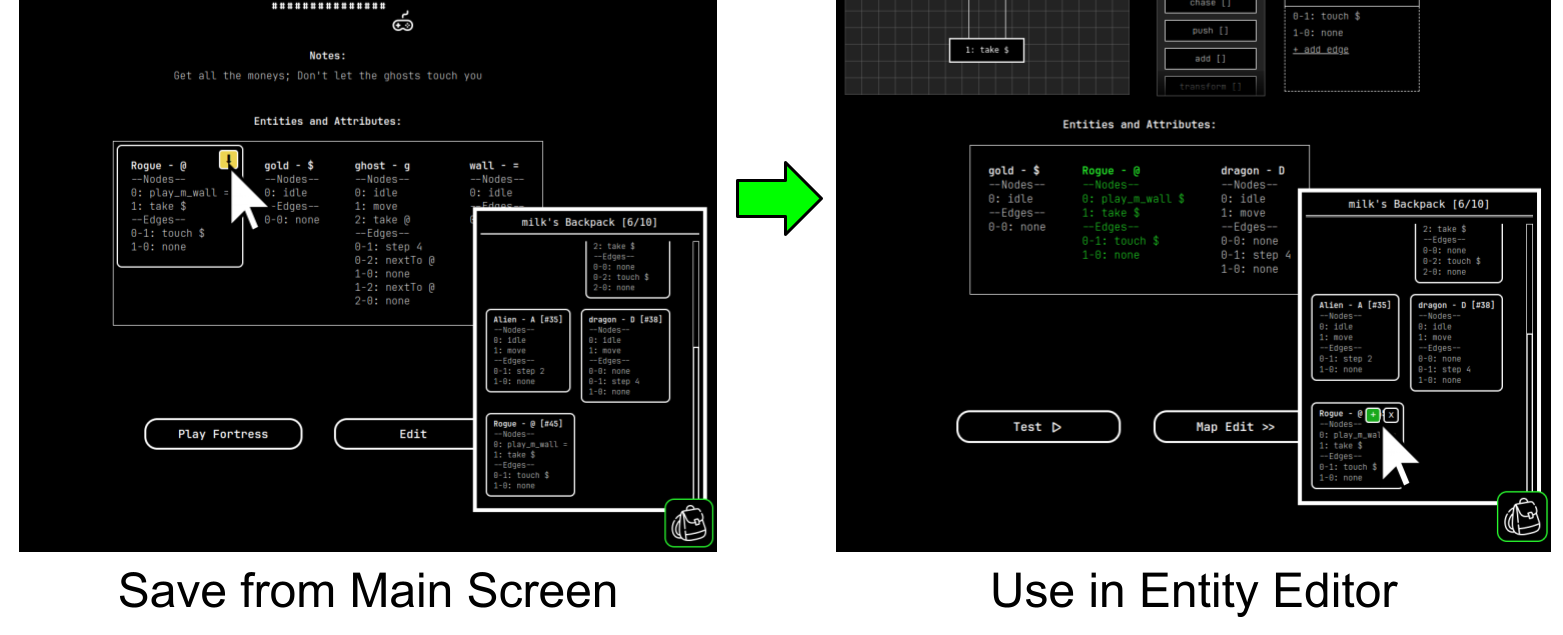}
    \caption{When logged in, a user can save an entity from one fortress to use in a new fortress with the backpack feature.}
    \label{fig:backpack}
\end{figure}

The backpack feature of Amorphous Fortress Online (demonstrated in Figure \ref{fig:backpack}) allows users to save entity characters from submitted fortresses to their user account. By clicking on an entity definition from the fortress view window of the main page, users can save entities to their "backpack" storage -- a pop-up window found at the bottom of the screen. 
Up to 10 entities can be saved to the user's "backpack" storage. These saved entities can then be added to a new fortress in the entity editor. If a backpack entity has a reference to a character not found in the fortress, it will be randomly replaced with another entity. For example, if a backpack entity \textit{A} contains the node \textit{take \$} but the fortress only contains entities with the characters \textit{\&, M, and +}, then the node will be changed on addition to the fortress to either \textit{take \&}, \textit{take M}, \textit{take +}, or \textit{take A}. 
This feature facilitates collaboration and remixing across environment design and the usage trends of different entity definitions and roles can be studied across fortress development. 

\subsection{User Guide}

A user guide\footnote{http://amorphous-fortress.xyz/wiki/} is provided for Amorphous Fortress Online as a tutorial, reference manual, and explanation for elements of the engine. The guide features sections on the Amorphous Fortress engine itself, the main screen, entity, fortress, and text editors, and the play screen. On each page of the guide are demonstrative GIFs to show the player how to interact with the Amorphous Fortress Online interface using examples from previously submitted fortresses. 


\section{Short Study}

We released the Amorphous Fortress Online v1 site on February 20, 2024. The website was promoted on various social media sites including X (formerly known as Twitter), BlueSky, LinkedIn, Facebook, and Reddit.



While we didn't receive many new fortress submissions, we collected a total of 174 plays from the fortresses. Six authors submitted a total of 33 fortresses to the website. Unfortunately, a majority of these fortresses come from the authors -- only five additional fortresses were submitted after the release of the site. This stark difference of fortress plays vs. fortress creation can possibly attribute to a large learning curve to using the system. An unfamiliarity with the engine as opposed to more familiar domains and games that have online editors such as Free Rider and Baba is You may have also attributed to the lack of submissions. 

\begin{figure}[!h]
    \centering
    \includegraphics[width=\linewidth]{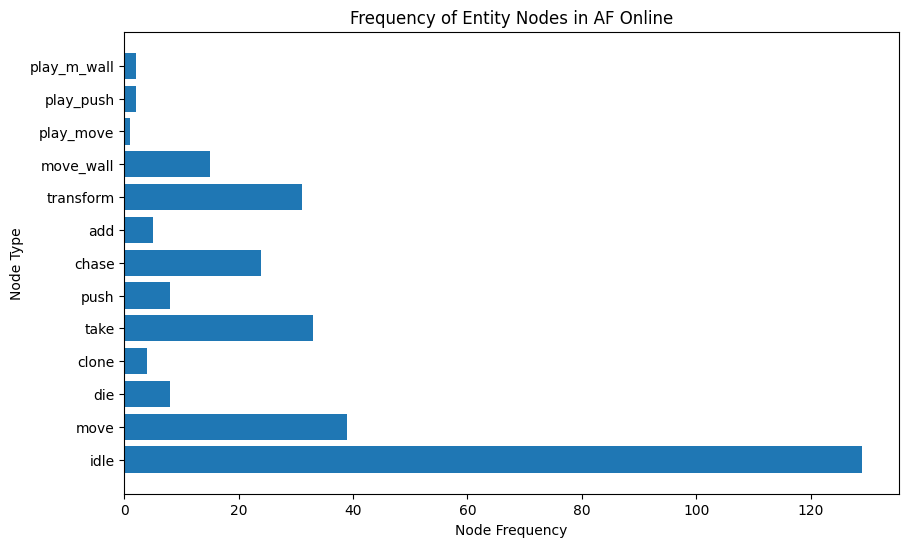}
    \caption{Frequency of entity action nodes}
    \label{fig:node_freq}
\end{figure}

Outside of the author's submission of 21 fortresses, users submitted an average of two fortresses. There were a total of 143 entities made in the website and a total of 298 nodes. Figure \ref{fig:node_freq} shows the frequency of each node type in the system. Interestingly, about 34.38\% of submitted fortresses were remixed, with the longest lineage of fortresses having a depth of 4 fortresses. Five fortresses with a \textit{play\_move}, \textit{play\_push}, or \textit{play\_move\_wall} were made.


\section{Reflection and Future Work}

Amorphous Fortress Online is intended to be a system to lay the foundation for future generative and recommendation game AI studies by utilizing the user submitted data. With this data, we would like to accentuate the environment design process and make the system more engaging by introducing mixed-initiative and collaborative AI systems trained from on-going submitted fortress data. The fortress information submitted to Amorphous Fortress Online along with the website source code will also be made open source and open access.

\subsection{Mixed-Initiative Fortress Design}
The lack of fortress submissions is one of the most pertinent areas we would like to address for future work. To remedy the intimidation factor of using the system and trying to make a new fortress from scratch or from a remix of another user's fortress, we will introduce the fortress evolver system used in the design of the first 2 iterations of the system \cite{charity2023amorphous,earle2023quality}. This generator will allow users to have a starting point and edit the fortress towards their liking -- much like the creative process in the Baba is Y'all system. This mixed-initiative interface will also encourage and develop an effective feedback system between the AI generator and the human designer to create novel fortress environments for the system.

\subsection{Entity Recommendation}

Designing and constructing thematically and semantic relevant entities for a fortress is a tedious process. This process is slightly alleviated with the backpack feature, however we would like to incorporate the entity and fortress-focused evolutionary components of the previous Amorphous Fortress works in the online version of the system to create a human-AI collaborative pipeline. Using the quality diversity methodology from previous work on the Amorphous Fortress \cite{earle2023quality}, we will build a recommendation system for the entity editor. This system will suggest thematically relevant entities to the user that also enhance the multi-agent interactions within the fortress. This recommendation system will be modeled after the variance of roles entities and NPCs have in video games, ranging from decoration entities and items to simple background characters that act as NPCs to more complex entities such as enemy characters. The complexity of these entity FSM definitions would complement the user's current entities in the fortress and guide the fortress towards emergent behavior for the environment.

\subsection{Sprite Generation}

The visuals of the Amorphous Fortress Engine are abstract and require more creative imagination from the user to interpret the actions and behaviors of the entities in the fortress. To make the system more visually appealing and engaging, the next iteration of Amorphous Fortress Online will feature a sprite editor to allow users to replace the entity ASCII character definitions with small 8x8 pixel 4-color sprites. The minimalist sprite designs is inspired by that of other easy-to-use microgame engines like PICO-8 and Bitsy. After users submit labeled text-sprite associations to their entities in the fortress, we will use this data to train a small generator model similar to architecture of the \$5 model \cite{merino2023five} to recommend and generate new sprites for the fortresses. This project will create a new database of labeled small scale images that will also continuously be updated based on site usage.





\section{Conclusion}

This paper introduces Amorphous Fortress Online -- an online simulation / game design system based around community-centric user generated environment creation. We describe the features and interfaces available to users to engage with the site and submit open-ended and thematically diverse games and multi-agent environments. The growth and addition of new fortresses will set the groundwork to build a large dataset that can be used for future AI research in agent training and generative models.

\bibliographystyle{ieeetr}
\bibliography{bibliography}

\end{document}